\documentclass[letterpaper, 10 pt, conference]{ieeeconf} 

\IEEEoverridecommandlockouts                        

\usepackage{graphics}
\usepackage{amsmath} 
\usepackage{amssymb}
\usepackage{cite}
\usepackage{tikz}
\usepackage{amsmath}
\usepackage{tabularx}
\usepackage{array}
\usepackage{booktabs}
\usepackage{bm}
\usepackage{algorithm}
\usepackage{algorithmic}
\usepackage{bbm}
\usepackage{hyperref}
\usepackage{units}
\usepackage{subfigure}
\usepackage{caption}
\usepackage{tablefootnote}

\definecolor{green}{RGB}{3,200,15}

\title{\LARGE \bf
Autonomous Ground Navigation in Highly Constrained Spaces: \\Lessons learned from The 2nd BARN Challenge at ICRA 2023
}
\author{\textbf{Competition Organizers}: Xuesu Xiao$^{1}$, Zifan Xu$^{2}$, Garrett Warnell$^{2, 3}$, and Peter Stone$^{2, 4}$, 
\\\textbf{Winning Teams}: Ferran Gebelli Guinjoan$^{5}$, R\^omulo T. Rodrigues$^{6}$, Herman Bruyninckx$^{6,7}$, \\ Hanjaya Mandala$^{8}$, Guilherme Christmann$^{8}$,\\  Jose Luis Blanco-Claraco$^{9}$, and Shravan Somashekara Rai$^{10}$
\thanks{$^{1}$George Mason University
$^{2}$The University of Texas at Austin
$^{3}$Army Research Laboratory
$^{4}$Sony AI 
$^{5}$Flanders Make
$^{6}$Flanders Make@KU Leuven
$^{7}$Eindhoven University of Technology
$^{8}$Inventec Corporation
$^{9}$University of Almeria
$^{10}$Teladoc Health Inc.
}
}
\begin{document}

\maketitle
\thispagestyle{empty}
\pagestyle{empty}

\begin{abstract}

The 2nd BARN (Benchmark Autonomous Robot Navigation) Challenge took place at the 2023 IEEE International Conference on Robotics and Automation (ICRA 2023) in London, UK and continued to evaluate the performance of state-of-the-art autonomous ground navigation systems in highly constrained environments. 
Compared to The 1st BARN Challenge at ICRA 2022 in Philadelphia, the competition has grown significantly in size, doubling the numbers of participants in both the simulation qualifier and physical finals: 
Ten teams from all over the world participated in the qualifying simulation competition, six of which were invited to compete with each other in three physical obstacle courses at the conference center in London, and three teams won the challenge by navigating a Clearpath Jackal robot from a predefined start to a goal with the shortest amount of time without colliding with any obstacle. 
The competition results, compared to last year, suggest that the teams are making progress toward more robust and efficient ground navigation systems that work out-of-the-box in many obstacle environments. However, a significant amount of fine-tuning is still needed onsite to cater to different difficult navigation scenarios. Furthermore, challenges still remain for many teams when facing extremely cluttered obstacles and increasing navigation speed.
In this article, we discuss the challenge, the approaches used by the three winning teams, and lessons learned to direct future research. 

\end{abstract}
\section{The 2nd BARN Challenge Overview}
\label{sec::challenge}

The 2nd BARN  (Benchmark Autonomous Robot Navigation) Challenge~\cite{the_2nd_barn_challenge} took place as a conference competition at the 2023 IEEE International Conference on Robotics and Automation (ICRA 2023) in London, UK. As a continuation of The 1st BARN Challenge at ICRA 2022 in Philadelphia,  the 2nd challenge aimed to evaluate the capability of state-of-the-art navigation systems to move robots through static, highly-constrained obstacle courses, an \emph{ostensibly} simple problem even for many experienced robotics researchers, but in fact, as the results from the 1st competition suggested, a problem far away from being solved~\cite{xiao2022autonomous}. 

Each team needed to develop an entire navigation software stack for a standardized and provided mobile robot, i.e., a Clearpath Jackal~\cite{clearpath_jackal} with a 2D 270\textdegree-field-of-view Hokuyo LiDAR for perception and a differential drive system with $2\textrm{m/s}$ maximal speed for actuation.
The developed navigation software stack needed to autonomously drive the robot from a given starting location through a dense obstacle field and to a given goal without any collisions with obstacles or any human interventions.
The team whose system could best accomplish this task within the least amount of time would win the competition.
The 2nd BARN Challenge had two phases: a qualifying phase evaluated in simulation, and a final phase evaluated in three physical obstacle courses.
The qualifying phase took place before the ICRA 2023 conference using the BARN dataset~\cite{perille2020benchmarking} (with the recent addition of DynaBARN~\cite{nair2022dynabarn}), which is composed of 300 obstacle courses in Gazebo simulation randomly generated by cellular automata.
The top six teams from the simulation phase were then invited to compete in three different physical obstacle courses set up by the organizers at ICRA 2023 in the ExCeL London conference center.

In this article, we report on the simulation qualifier and physical finals of The 2nd BARN Challenge at ICRA 2023, present the approaches used by the top three teams, discuss lessons learned from the challenge compared against The 1st BARN Challenge at ICRA 2022, and point out future research directions to solve the problem of autonomous ground navigation in highly constrained spaces. 
\section{Simulation Qualifier}
\label{sec::simulation}
\begin{figure*}[t]
    \centering
    \includegraphics[width=2\columnwidth]{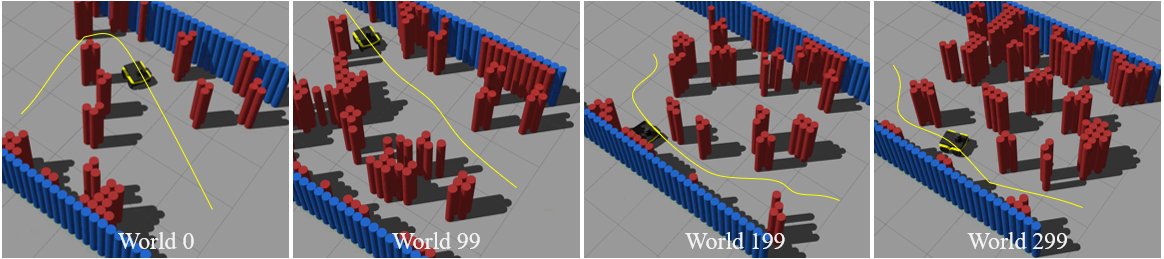}
    \caption{Four example BARN environments in the Gazebo simulator (ordered by ascending relative difficulty level).}.
    \label{fig::barn_worlds}
\end{figure*}

The simulation qualifier of The 2nd BARN Challenge started on January 1\textsuperscript{st}, 2023. 
The qualifier used the BARN dataset~\cite{perille2020benchmarking}, which consists of 300 $5\textrm{m}\times5\textrm{m}$ obstacle environments randomly generated by cellular automata (see examples in Fig.~\ref{fig::barn_worlds}), each with a predefined start and goal. 
These obstacle environments range from relatively open spaces, where the robot barely needs to turn, to highly dense fields, where the robot needs to squeeze between obstacles with minimal clearance.
The BARN environments are open to the public, and were intended to be used by the participating teams to develop their navigation stack.
Another 50 unseen environments, which are not available to the public, were generated to evaluate the teams' systems.
A random BARN environment generator was also provided to the teams so that they could generate their own unseen test environments.\footnote{\url{https://github.com/dperille/jackal-map-creation}}

In addition to the 300 BARN environments, six baseline approaches were also provided for the participants' reference, ranging from classical sampling-based~\cite{fox1997dynamic} and optimization-based navigation systems~\cite{quinlan1993elastic}, to end-to-end machine learning methods~\cite{xubenchmarking, wang2021agile}, and hybrid approaches~\cite{xu2021applr}.
All baselines were implementations of different local planners used in conjunction with Dijkstra's search as the global planner in the ROS \texttt{move\_base} navigation stack~\cite{rosmovebase}.
To facilitate participation, a training pipeline capable of running the standardized Jackal robot in the Gazebo simulator with ROS Melodic (in Ubuntu 18.04), with the option of being containerized in Docker or Singularity containers for fast and standardized setup and evaluation, was also provided.\footnote{\url{https://github.com/Daffan/ros_jackal}}

\subsection{Rules}
Each participating team was required to submit their developed navigation system as a (collection of) launchable ROS node(s).
The challenge utilized a standardized evaluation pipeline\footnote{\url{https://github.com/Daffan/nav-competition-icra2022}} to run each team's navigation system and compute a standardized performance metric that considers navigation success rate (collision or not reaching the goal counts as failure), actual traversal time, and environment difficulty (measured by optimal traversal time).
Specially, the score $s$ for navigating each environment $i$ was computed as
\[
s_i = 1^{\textrm{success}}_i \times \frac{\textrm{OT}_i}{\textrm{clip}(\textrm{AT}_i, 4\textrm{OT}_i, 8\textrm{OT}_i)} \; ,
\]
where the indicator function $1_\textrm{success}$ evaluates to $1$ if the robot reaches the navigation goal without any collisions, and evaluates to $0$ otherwise.
$\textrm{AT}$ denotes the actual traversal time, while $\textrm{OT}$ denotes the optimal traversal time, as an indicator of the environment difficulty and measured by the shortest traversal time assuming the robot always travels at its maximal speed ($2\textrm{m/s}$):
\[
\textrm{OT}_i = \frac{\textrm{Path Length}_i}{\textrm{Maximal Speed}}.
\]
The Path Length is provided by the BARN dataset based on Dijkstra's search from the given start to goal.
The $\textrm{clip}$ function clips $\textrm{AT}$ within 4OT and 8OT in order to assure navigating extremely quickly or slowly in easy or difficult environments respectively won't disproportionally scale the score. 
Note that the hyper-parameters $4$ and $8$ for $\textrm{OT}$ are manually selected before The 1st BARN Challenge, and the organizers found out that the performance of the submitted navigation systems in The 2nd BARN Challenge has reached the upper bound of this specific metric, $0.25$. So the organizers plan to change these hyper-parameters next year to increase the upper bound. 
The overall score of each team is the score averaged over all 50 unseen test BARN environments, with 10 trials in each environment. Higher scores indicate better navigation performance.
The six baselines score between $0.1627$ and $0.2334$~\cite{the_2nd_barn_challenge}.

\subsection{Results}
The simulation qualifier started on January 1\textsuperscript{st}, 2023 and lasted through a soft submission deadline (April 20\textsuperscript{th}, 2023) and a hard submission deadline (May 20\textsuperscript{th}, 2023). Submitting by the soft deadline will guarantee an invitation to the final physical competition given good navigation performance in simulation and leave sufficient time for invited participants to make travel arrangements to London. The hard deadline is to encourage broader participation, but final physical competition eligibility will depend on the available capacity and travel arrangement made beforehand.  
In total, ten teams from all over the world submitted their navigation systems.
The performance of each submission was evaluated by the standard evaluation pipeline. The results are shown in Tab.~\ref{tab::sim_results} with the baselines shown in the forth column as a reference. 

\begin{table}[h]
  \caption{Simulation Results}
  \label{tab::sim_results}
  \centering
  \small
  \begin{tabular}{cccc}
  \toprule
  Rank. & Team & Score & Baseline \\
  \midrule
  1 & KUL+FM & 0.2490\\
  2 & INVENTEC & 0.2445\\
  3 & University of Almeria & 0.2439\\
  4 & UT AMRL & 0.2424 & LfLH~\cite{wang2021agile} \\
  5 & Temple TRAIL & 0.2290\\
  6 & UVA AMR & 0.2237\\
  7 & RIL & 0.2203 & E-Band~\cite{quinlan1993elastic}, e2e~\cite{xubenchmarking}\\
  8 & Staxel & 0.2019 & APPLR-DWA~\cite{xu2021applr}\\
  9 & The MECO Barners & 0.1829 & (Fast) DWA~\cite{fox1997dynamic}\\
  10 & Lumnonicity & NA\\
  \bottomrule
  \end{tabular}
\end{table}

Compared to the simulation competition in The 1st BARN Challenge at ICRA 2022, in which only one team (Temple TRAIL) outperformed all baselines, four teams (KUL+FM, INVENTEC, University of Almeria, and UT AMRL) achieved better score than the best baseline, Learning from Learned Hallucination (LfLH, 0.2334)~\cite{wang2021agile}. The top six teams, KUL+FM (KU Leuven and Flanders Make), INVENTEC (Inventec Corporation), University of Almeria (University of Almeria and Teladoc Health Inc.), UT AMRL (The University of Texas at Austin), Temple TRAIL (Temple University), UVA AMR (University of Virginia) were invited to the physical finals at ICRA 2023. Note that Temple TRAIL's simulation score was decreased compared to last year, since the team has learned from last year's experience and focused on the sim-to-real gap. 

\section{Physical Finals}
\label{sec::physical}
\begin{figure}[ht]
    \centering
    \includegraphics[width=1\columnwidth]{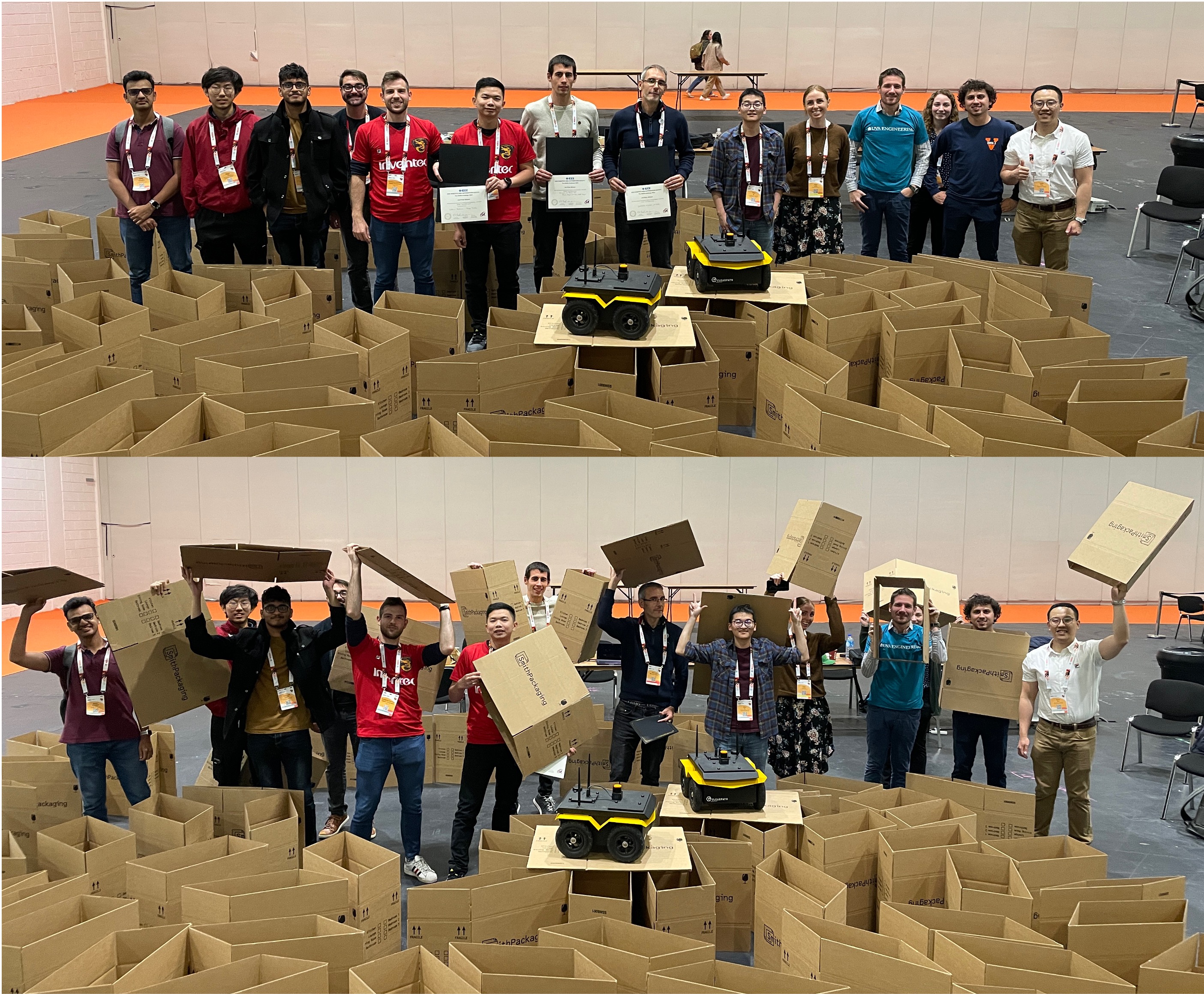}
    \caption{Final physical competition participants, sponsor Clearpath Robotics, and organizers at The 2nd BARN Challenge in London, UK. }
    \label{fig::barn_at_icra}
\end{figure}

The physical finals took place at ICRA 2023 in the ExCel London Conference Center on May 30\textsuperscript{th} and May 31\textsuperscript{st}, 2023 (Fig.~\ref{fig::barn_at_icra}).
Two physical Jackal robots with the same sensors and actuators were provided by the competition sponsor, Clearpath Robotics. 

\subsection{Rules}
\begin{figure}
    \centering
    \includegraphics[width=1\columnwidth]{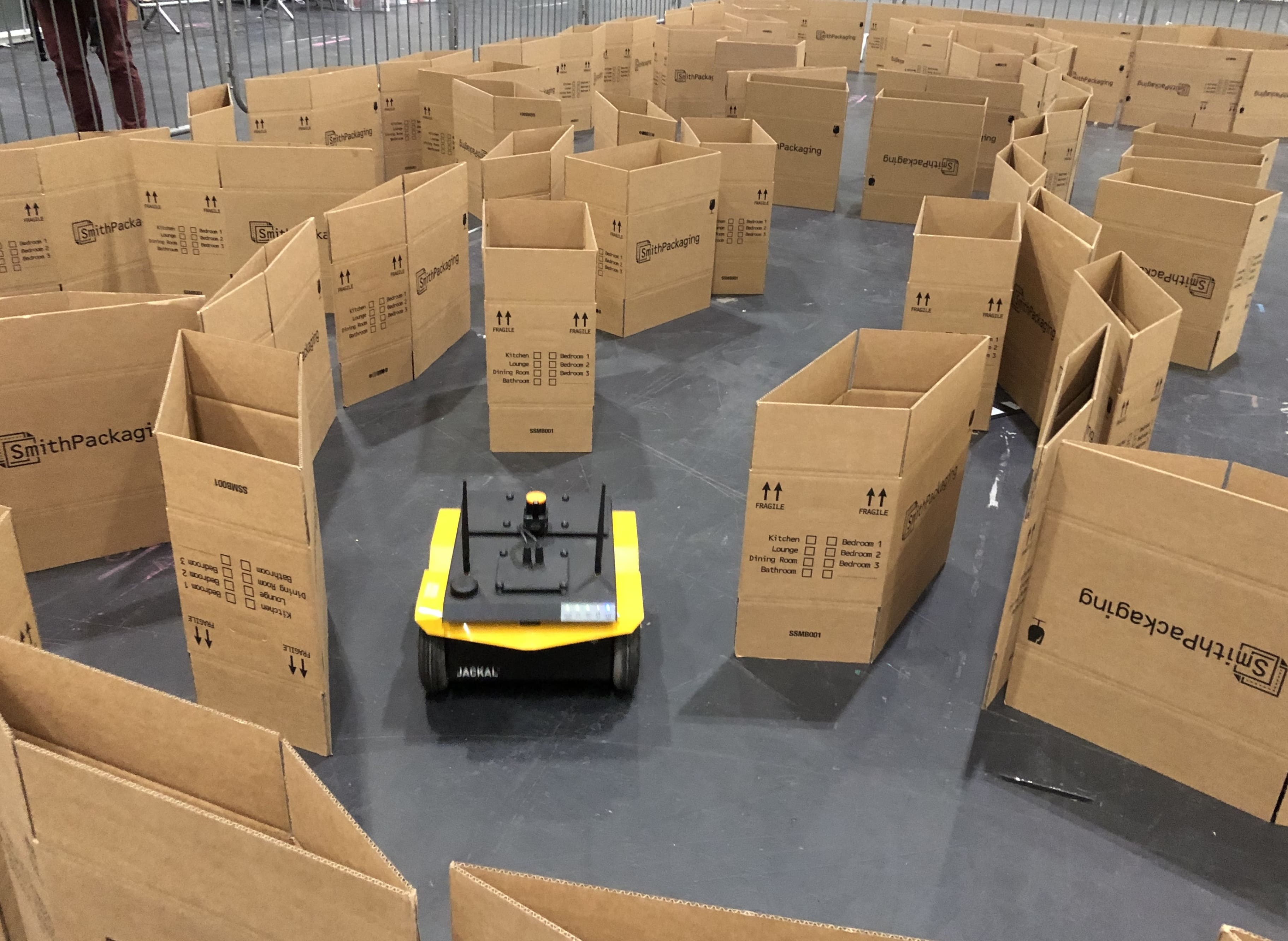}
    \caption{One (out of three) physical obstacle courses during the finals.}
    \label{fig::physical_course}
\end{figure}

Physical obstacle courses were set up using 90 cardboard boxes in the conference center (Fig.~\ref{fig::physical_course}).
The organizers used the same guidelines to set up three obstacle courses as in The 1st BARN Challenge, i.e., all courses aimed at testing a navigation system's local planning and therefore had an obvious passage but with minimal clearance (a few centimeters around the robot) when traversing this passage. 

The organizers also used the same competition rules agreed by all the physical competition participants: Although it was still impractical to run exactly the same navigation systems submitted by the teams in the simulation qualifier due to poor performance in the real world, the organizers reduced both the set-up time and competition time from last year's 30 minutes to 20 minutes. After the 20-minute set-up time, each team had the opportunity to run five timed trials (after notifying the organizers to be timed) within another 20-minute. The fastest three out of the five timed trials were counted, and the team that had the most successful trials (reaching the goal without any collision) was the winner.
In the case of a tie, the team with the fasted average traversal time would be declared the winner. 

\subsection{Results}
The six teams' navigation performance is shown in Tab.~\ref{tab::physical_results}. Note that due to travel-related reasons, Temple TRAIL cannot attend in-person, so the organizers ran their system submitted to the simulation qualifier for them, which, unfortunately but also as expected, cannot finish one single trial without onsite fine-tuning. 
Compared to last year, many teams struggled less on the obstacle avoidance problem in the first two easier environments, and therefore were able to shift their attention to increasing speed and were mostly navigating at a much higher speed ($>1.0\textrm{m/s}$). The detailed results of all five timed trials (only the top three were counted in the final score) are listed in the last three columns of Tab.~\ref{tab::physical_results}, where ``X'' indicates failure. 

The winner, KUL+FM, is the very first team in The BARN Challenge history that has finished all nine counted physical trials without any collision. In fact, they only failed three trials in all 15 timed trials in the three obstacle courses. The 2nd place winner, INVENTEC, was able to quickly navigate all six counted trials in the first two environments, sometimes even faster than KUL+FM, but didn't manage to finish the last most constrained obstacle course. University of Almeria also failed all three trials in the last course and one in the first one. 

\begin{table*}[ht]
  \caption{Physical Results}
  \label{tab::physical_results}
  \centering
  \small
  \begin{tabular}{ccccccc}
  \toprule
  Rank. & Team & Success/Total & Average Time & Course 1 & Course 2 & Course 3 \\
  \midrule
  1 & KUL+FM & 9/9 & 34/56/91 & 36/X/33/34/34 & 63/64/52/47/54 & 86/79/X/79/X\\
  2 & INVENTEC & 6/9 & 44/64/NA & 47/42/45/45/41 & 58/66/X/67/X & X/X/X/X/X\\
  3 & University of Almeria & 5/9 & 119/79/NA & X/103/134/X/X & 85/86/93/X/53 & X/X/X/X/X\\
  4 & UVA & 4/9 & 68/NA/103 & X/X/76/57/71 & X/X/X/X/X & 103/X/X/X/X \\
  5 & UT AMRL & 3/9 & 84/NA/NA & 91/X/79/X/81 & X/X/X/X/X & X/X/X/X/X  \\
  6 & Temple TRAIL & 0/9 & NA/NA/NA & X/X/X/X/X & X/X/X/X/X & X/X/X/X/X \\
  \bottomrule
  \end{tabular}
\end{table*}

\section{Top Three Teams and Approaches}
\label{sec::teams}
In this section, we report the approaches used by the three winning teams.

\subsection{KU Leuven and Flanders Make (KUL+FM)}
\begin{figure}[ht!]
\centering
\subfigure[]{\label{fig:kul_fm_motion_tube_inflation}\includegraphics[width=.23\textwidth]{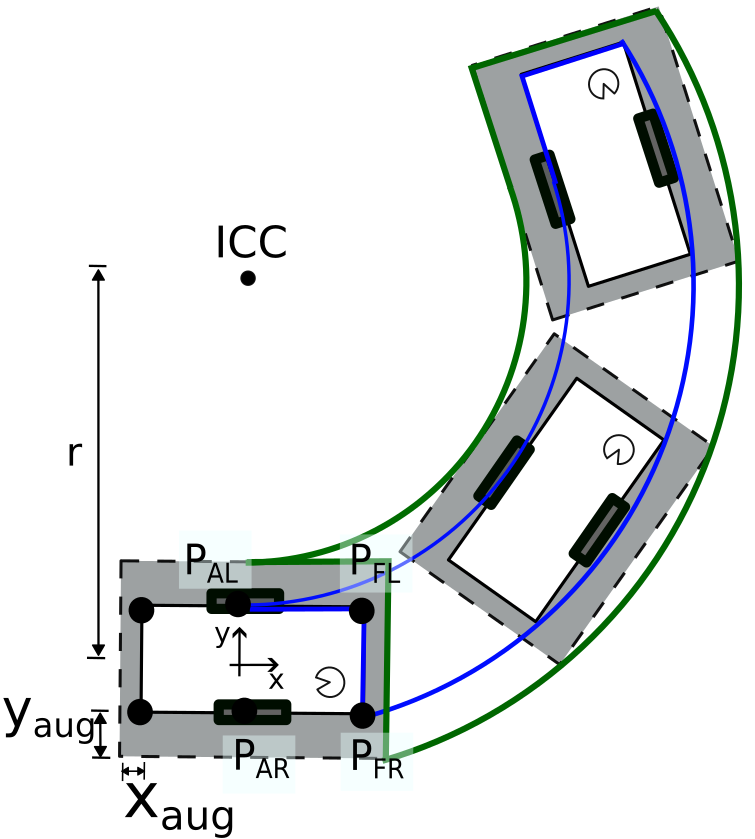}}
\subfigure[]{\label{fig:kul_fm_motion_tube_sampling}\includegraphics[width=.23\textwidth]{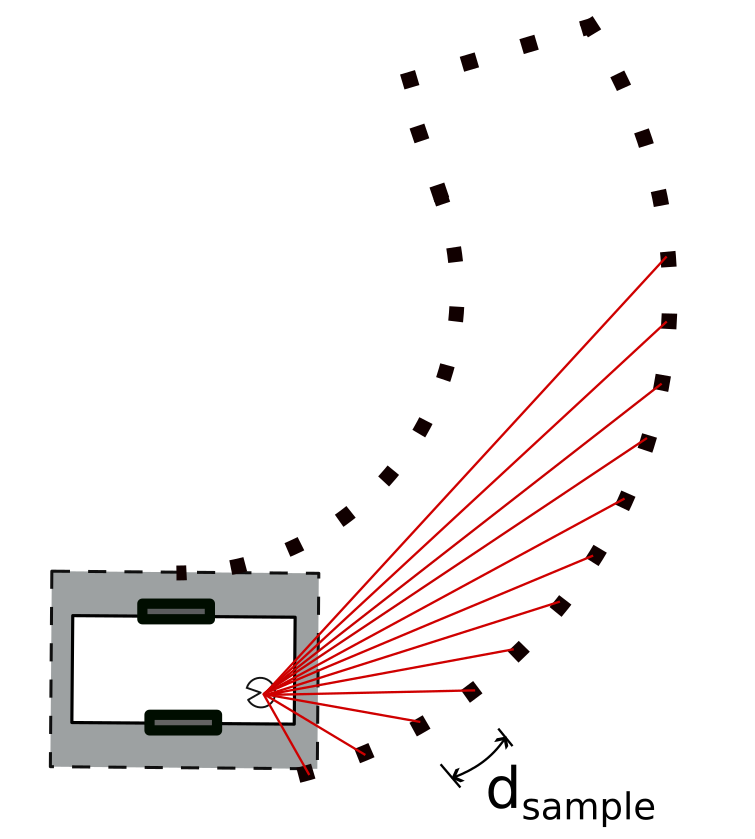}}
\caption{(KUL+FM) (a) A motion tube corresponds to the inflated trajectory of the vehicle. (b) For computational efficiency, samples are projected in the sensor space.}
\end{figure}

The core algorithm of the KUL+FM team is the adaptive free-space motion tube~\cite{RTHB23}. Consider that a robot's maneuver is defined by a curvature that the robot follows for a time horizon ($T$) at a constant forward velocity. The motion tube corresponds to the swept volume, that is, the area to be occupied by the vehicle when performing a maneuver. To deal with the discrete resolution of range sensors as well as measurement uncertainty, the footprint of the platform is inflated. Figure~\ref{fig:kul_fm_motion_tube_inflation} shows the motion tube for a particular maneuver using the physical (blue polygon) and the inflated (green polygon) footprint of the vehicle. To evaluate whether a motion-tube is within the free-space, the edge of the inflated swept volume is sampled at an interval $d_{sample}$ (Fig.~\ref{fig:kul_fm_motion_tube_sampling}). For computational efficiency, samples are projected in the sensor space by associating them to beam indices. A candidate maneuver is said to be available if, for each sample, the corresponding measurement reported by the sensor is greater than the distance from the sensor to the sample. The reasoning for choosing the inflation and sampling interval values are discussed in the original paper~\cite{RTHB23}. Finally, within the available motion tube, one can decide the control input of the vehicle based on a high-level application goal. 

In essence, instead of inflating obstacles in a local map, the adaptive motion tube inflates the robot and its corresponding trajectory. Because motion-tubes are computed with respected to the robot, there is no dependency between localization accuracy and free-space navigation. Another important characteristic of the method is the computational efficiency thanks to the projection of Cartesian samples in the sensor space, which allows computing thousands of motion tubes in low-end computers. A potential disadvantage is that the method relies on the current sensor reading, therefore it is not robust to outliers and measurement errors. Fortunately, both measurement outliers and errors are rather rare and negligible in the proximity of the sensor.

\textbf{Software stack:} The KUL+FM's software stack for The BARN Challenge uses adaptive free-space motion tube for local navigation, ROS Global Planner for global planning, and Hector SLAM~\cite{KMSK2011} for online mapping and tracking. Based on the most updated map, the global planner provides sub-goal for the local navigation. These sub-goals are used to assign costs to available motion tubes: the closer a maneuver takes the vehicle to a sub-goal, the smaller its cost is. The control input sent to the vehicle corresponds to the weighted average of the available tubes.

\textbf{Parameter tuning:} The parameter of the local navigation were kept constant during the simulation and the physical competition (2000 motion tubes with different curvatures and forward velocities, $T=1$~s, $d_{sample}=0.02$~m). Most of the parameter tuning took place in the global navigation: to avoid obstacles becoming too large and causing the global planner to fail in narrow passages, the team incrementally decreased the obstacle inflation and the distance of the obstacles to be considered in the planner. 

\subsection{INVENTEC}
The INVENTEC team's approach was to extend the best-performing baseline, Learning from Learned Hallucination (LfLH)~\cite{wang2021agile} with improved collision check and recovery behaviors via a Finite State Machine (FSM). The main driving mode relies on a learning-based model that learns to drive a robot by collecting random trajectories and hallucinating obstacles~\cite{xiao2021toward, xiao2021agile, wang2021agile}. However, to address poor generalization of the learned model in out-of-distribution deployment scenarios, the team introduced two alternative modules for front safety checks: Footprint Inflation (FI) used in the simulation qualifier and Model Predictive Control (MPC) used in the physical finals. Additionally, during backward movements, the team performed safety checks by extracting a Region of Interest (RoI) from the obstacle costmap in the robot memory. Details can be found in the comprehensive technical report by the INVENTEC team~\cite{mandala2023barn}. 

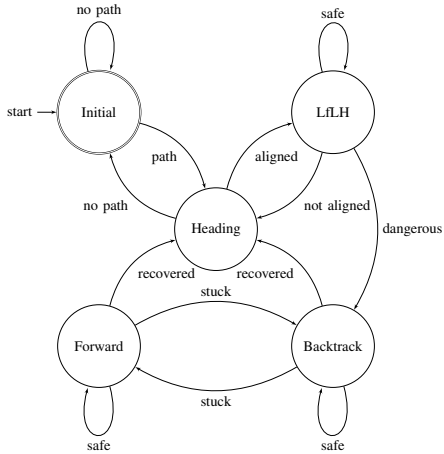
\begin{figure}[!htb]
    \scalebox{0.55}{\usetikzlibrary{arrows, automata}

\begin{tikzpicture}[node distance=4cm,auto,>=latex']
    \centering
    \node[initial,state, circle, accepting, minimum size=2.0cm]       (initial)     {Initial};
    \node[state, circle, minimum size=2.0cm]  (heading)   [below right of=initial]  {Heading};
    \node[state, circle, minimum size=2.0cm]  (LfLH)      [above right of=heading]  {LfLH};
    \node[state, circle, minimum size=2.0cm]  (backtrack) [below right of=heading]  {Backtrack};
    \node[state, circle, minimum size=2.0cm]  (forward)   [below left of=heading]   {Forward};

    \path[->] (initial)   edge [loop above]   node {no path}       (initial)
              (initial)   edge [bend left]    node[below left] {path}          (heading)
              (heading)   edge [bend left]    node {no path}       (initial)
              (heading)   edge [bend left]    node[below right] {aligned}       (LfLH)
              (LfLH)      edge [loop above]   node {safe}          (LfLH)
              (LfLH)      edge [bend left]    node {not aligned}   (heading)
              (LfLH)      edge [bend left]    node {dangerous}     (backtrack)
              (backtrack) edge [loop below]   node {safe}          (backtrack)
              (backtrack) edge [bend left]    node {stuck}         (forward)
              (backtrack) edge [bend right]   node {recovered}     (heading)
              (forward)   edge [loop below]   node {safe}          (forward)
              (forward)   edge [bend left]    node [below right] {recovered}     (heading)
              (forward)   edge [bend left]    node {stuck}         (backtrack);
\end{tikzpicture}}
    \caption{(INVENTEC) Navigation Finite State Machine.}
    \label{fig:inventec_fsm}
\end{figure}

\textbf{Navigation FSM:} The FSM consists of five states: \textit{Initial, Heading, LfLH, Forward, and Backtrack}, as shown in Fig.~\ref{fig:inventec_fsm}. In the \textit{Initial} state, the navigation controller waits for the path computed by Dijkstra’s search in the \texttt{move\_base} global planner with \textit{NavFn} plugin. Then, the state is switched to \textit{Heading}, which aligns the robot to the target path within a tolerance of $\pm 30^{\circ}$. Then, the \textit{LfLH} model produces velocity commands taking as input the current LiDAR scans and a local goal drawn from the global path 0.5m ahead of the robot. At every step a safety check is performed. If a future collision is detected, the state changes to \textit{Backtrack} recovery behavior and an alternative slow \textit{Forward} state if the robot is stuck during backtrack.

\begin{figure}[h]
    \centering
    \subfigure[]{\label{fig:jackal_costmap}\includegraphics[width=.175\textwidth]{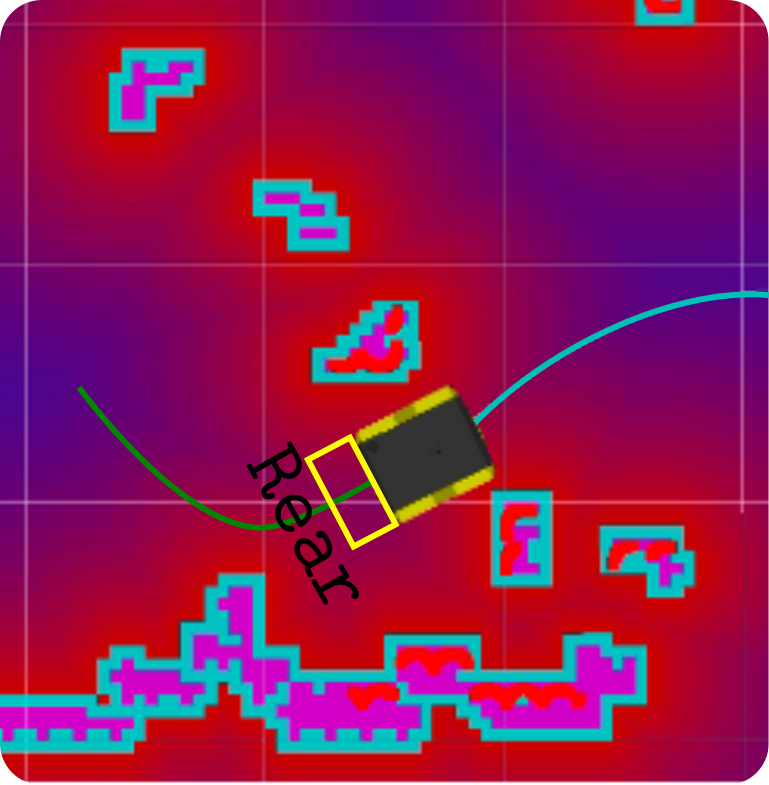}}
    \hspace{10mm}
    \subfigure[]{\label{fig:lidar_fov}\includegraphics[width=.175\textwidth]{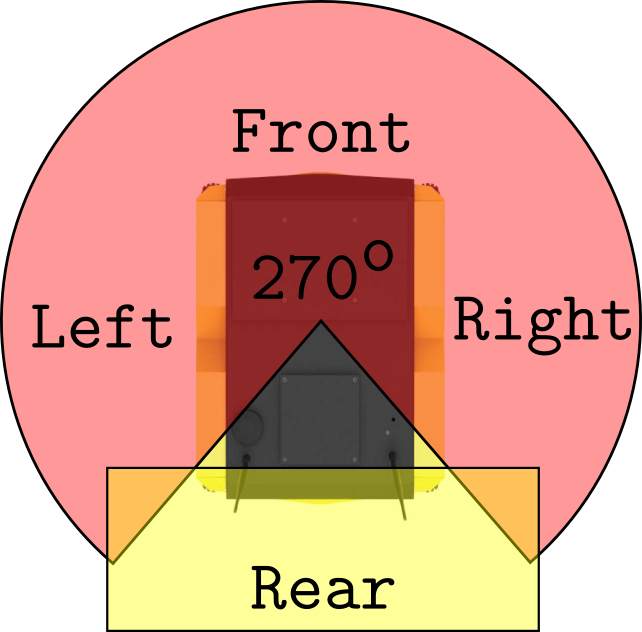}}
    \caption{(INVENTEC) (a) Costmap with robot rear RoI (yellow rectangle). (b) Jackal LiDAR field-of-view. }
\end{figure}

\textbf{Recovery Behaviors:} During the forward movement in \textit{LfLH} the robot's path is recorded (green line shown in Fig.\ref{fig:jackal_costmap}). When \textit{Backtrack} is first triggered, it samples a point 0.3 meters behind the robot along the recorded path, aligns the heading to the target point and performs a straight backward command. Moving backward means moving towards the LiDAR blind spot. Therefore, the team defined a rectangular RoI in the costmap directly behind the robot, illustrated in Fig.~\ref{fig:lidar_fov}). At every step, the method checks for possible collisions in the costmap RoI which contains information about past obstacles. If a potential collision is detected during the reverse movement, the state is switched to a slow-forward recovery behavior.

\begin{figure}[ht]
    \centering
    \includegraphics[width=0.7\columnwidth]{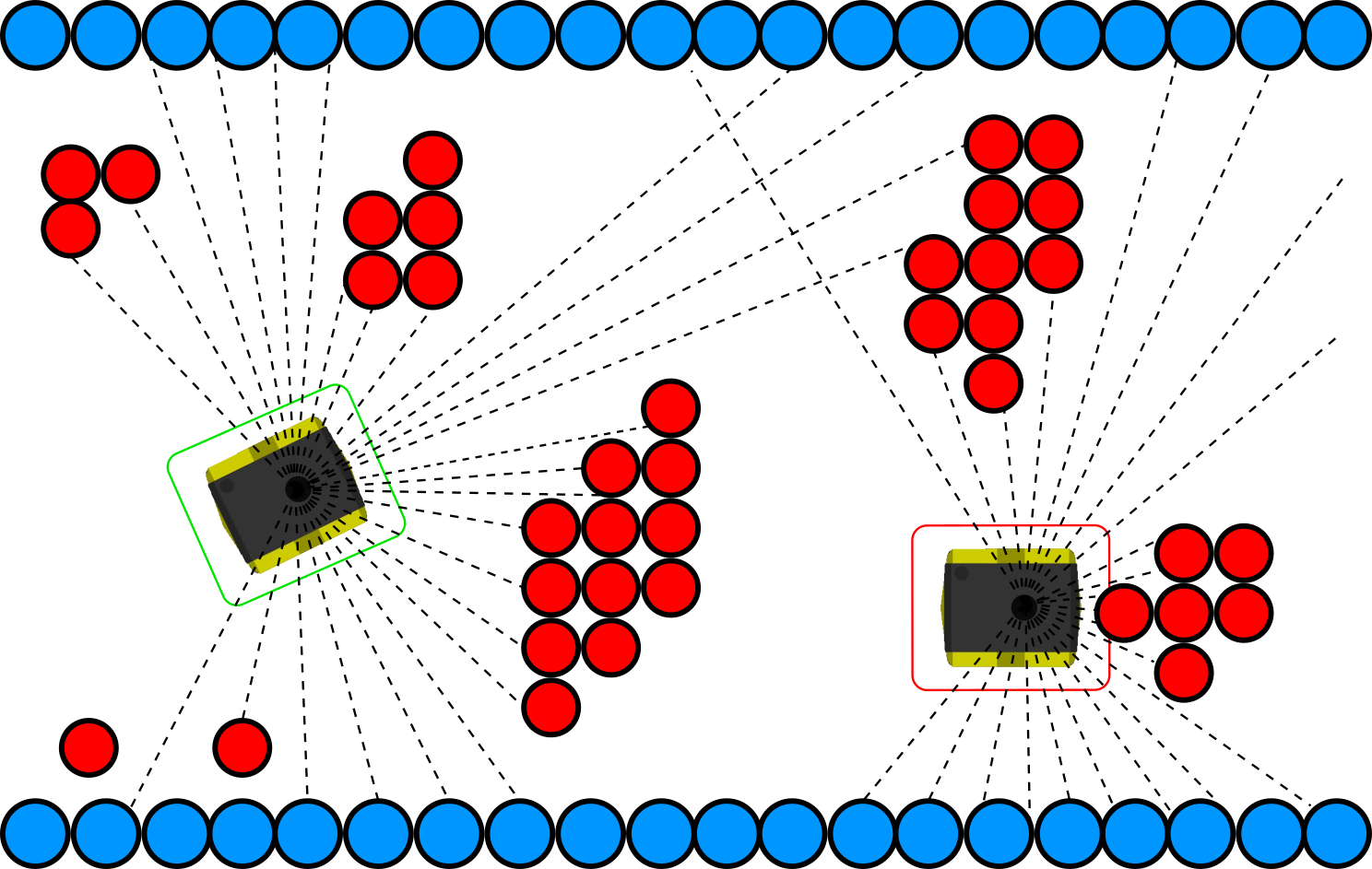}
    \caption{(INVENTEC) Footprint inflation for collision detection.}
    \label{fig:inventec_inflation}
\end{figure}

\textbf{Simulation Approach:} In the simulation stage, INVENTEC team's strategy for improving the baseline LfLH model was threefold. First, utilizing FI for obstacle checking with the latest LiDAR data (shown in Fig.~\ref{fig:inventec_inflation}); Second, checking the costmap for history obstacles to assess the safety of the robot's rear side; Finally, clipping maximum velocity to $0.7\textrm{m/s}$, which provided a significant performance improvement. The footprint inflation was $0.04$ m. As a result, whenever the inflated footprint overlaps any detected obstacles, the robot transitions into recovery behaviors. Two illustrations of the inflated footprint are depicted in Fig.~\ref{fig:inventec_inflation}, in which the green region and red region indicate safe and unsafe conditions respectively. 

During the simulation qualifier, this method enabled the robot to navigate in close proximity to obstacles without collision. However, there exists a trade-off between the size of the inflated footprint and the maximum velocity. If the size is large, the robot will have enough time for braking at high speed but is unable to navigate in highly constrained environments as it frequently stops. Conversely, if the inflated footprint size is small, there is a higher probability of front collision due to insufficient time for braking. The team conducted a test and determined that a $0.7\textrm{m/s}$ maximum velocity paired with a $0.04m$ offset provided the optimal balance, allowing the robot to stably navigate without encountering a front collision.

\begin{figure}[h]
    \centering
    \includegraphics[width=0.7\columnwidth]{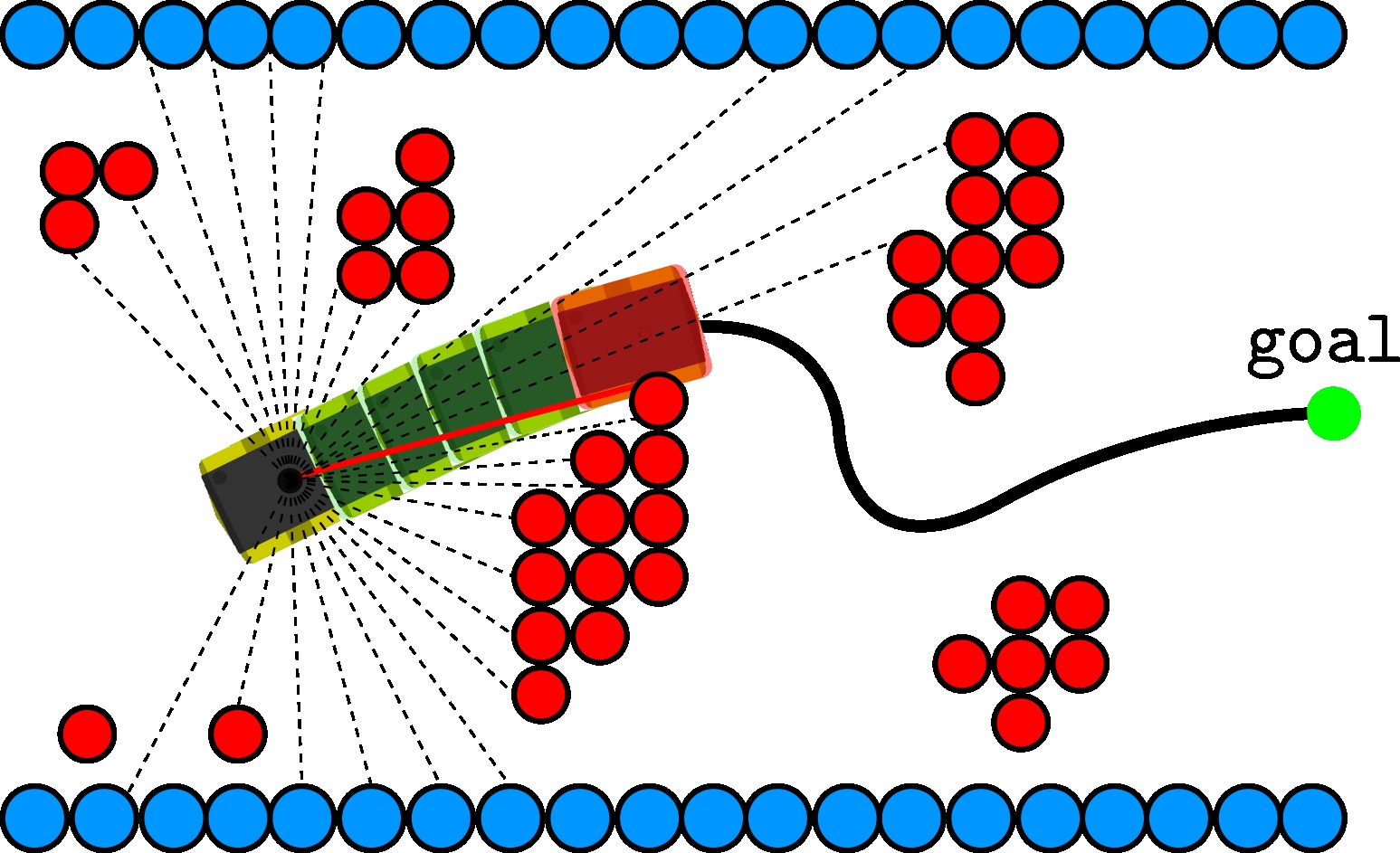}
    \caption{(INVENTEC) MPC footprint forward safety check based on LiDAR sensor.}
    \label{fig:inventec_mpc}
\end{figure}

\textbf{Real-world Approach: } The only difference between simulation and real-world methods was in the safety check when moving forward. A Model Predictive Control (MPC) approach was adopted for the safety check during forward movement in the real world. MPC takes into account not only the robot's current footprint but also integrates future position information, enabling the robot to proactively plan its movement and adjust its trajectory accordingly. MPC's suitability for predictive collision results in adaptability and real-time responsiveness in real-world navigation.

During the \textit{LfLH} operation, the MPC predicts future steps and detects collisions. At every step, it predicts 20 steps into the future (taking about 200 milliseconds). If at any point in the future trajectory an obstacle overlaps with the footprint of the robot, the current command is deemed unsafe. This stops the LfLH state and the FSM switches to recovery behaviors. Fig.~\ref{fig:inventec_mpc} depicts this process, with green indicating safe portions of the trajectory and red showing a detected future collision. 

\subsection{University of Almeria and Teladoc Health}

The University of Almeria team's implementation was built upon the Mobile Robots Programming Toolkit (MRPT)~\cite{blanco2014mobile}, an open-source C++ framework specifically developed for robotics applications, including libraries for navigation. The MRPT ROS nodes were fine-tuned to align with the specifications of the Clearpath Robotics Jackal robot~\cite{clearpath_jackal}.

A notable difference between the simulation qualifier and physical finals is that in simulation the goal was well-known, with fixed coordinates in a map known a priori. In contrast, in the physical finals there were no absolute coordinates of the goal. In practice, this makes the real-world navigation close to pure \emph{exploration}.

Those caveats aside, the architecture of the system comprises the subsystems enumerated in the following paragraphs, each one implemented as an independent ROS node. Most of the nodes were
used unmodified with respect to their open source repositories.\footnote{\url{https://github.com/mrpt-ros-pkg/}}

\textbf{Localization:} A custom implementation of particle filter-based localization \cite{thrun2001robust} with adaptive number of 
particles using the KLD approach \cite{fox2001kld} has been used by the University of Almeria team. 
This package is similar to the standard \texttt{amcl} ROS package,
but capable of using several metric maps at once for localization, 
with more than one implementation of particle filter algorithms \cite{blanco2010optimal}.

\textbf{Local obstacles map:} The purpose of this node is to perform real-time acquisition and processing of sensor data, in this case the LiDAR scans. This raw sensor data was subsequently transformed into a 2D point cloud representation, localized within the robot's own coordinate frame and decimated into lower resolution for faster processing. Phony points were also added in the real-world phase
along the past robot poses within a certain time window, in an 
attempt to discourage the navigation system to go back and revisit parts of the environment that were already traversed. 
This was revealed to be important when going through large open spaces, 
where turning back becomes possible.

\textbf{Path planner (higher layer):} At a relatively low rate ($1\textrm{Hz}$ or slower), the local obstacles were considered to find a kinematically feasible path using a path planning algorithm, which is then sent to a local path follower. The team's path planner uses a custom algorithm based on A* on a discrete lattice of
the state-vector space of the vehicle, i.e., SE(2) pose plus velocities.
Arcs between the lattice nodes were explored efficiently 
using Parameterized Trajectory Generators (PTGs), a concept derived from past works~\cite{Blanco2008}, which defines families of 
paths to help exploring the environment with kinematically and dynamically feasible paths.
Disregarding the use of these path libraries,
the team's method has resemblances with the implementation of the dubbed \emph{Smac Planner} in Nav2 \cite{macenski2020marathon2}, although a systematic comparison with the state-of-the-art 
and a proper paper reporting the method's details still remain as future work.
The whole planner can be seen as an improved version of a formerly published RRT planner \cite{blanco2015rrt}. This package is open-sourced and released to ROS.\footnote{\url{https://github.com/MRPT/mrpt_path_planning/}} 
This path planner is one of the main strengths of the whole stack, especially for simulation, since complex maneuvers in tight spaces can be computed in a safe manner.

\textbf{Local planner (lower layer):} Once a path is found reaching (or, at least, approaching) the target, the task of generating
motor commands to follow it, including avoiding any new obstacles, is accomplished by a reactive navigation system, 
which is also based on Trajectory Parameter Space (TP-Space) \cite{Blanco2008}.
In TP-Space, the robot, regardless of its physical shape and kinematic constraints, is transformed into a free-flying point within a newly formulated parameter space. This transformation incorporates the robot's shape and kinematic restrictions, thereby allowing for efficient navigation by taking into account the robot's specific physical characteristics and movement capabilities. On the other hand, PTGs define a set of potential trajectories for the robot, parameterized by variables such as path shape, robot speed, and turning radius. By dynamically adjusting these parameters based on real-time sensor data, the robot can select the optimal path for avoiding obstacles and reaching its destination.
The reactive navigation system, through its integration with TP-Spaces and PTGs, acts as an advanced behavior planning algorithm. The implementation used the ROS node developed based on these principles. It processed the 2D point cloud data to dynamically generate an optimal path for the robot, while simultaneously accounting for the robot's kinematic constraints and potential obstacles within its environment.
This local planner can run at a higher rate than the path planner,
typically between $5$ and $10\textrm{Hz}$.

\textbf{Exploration:} This module is in charge of generating targets for the path planner to find empty passages around.
Since The BARN Challenge typically requires the robot to ``move forward'', The University of Almeria team's implementation simply generated target goals in a region of interest a few meters in front of the robot, but with a certain random perturbation to avoid getting the global planner trapped in a local minima.

In summary, the system's ability to process and respond to real-time sensor data in a rapid and efficient manner, while accounting for the robot's physical and kinematic characteristics, makes it a robust solution for navigating complex and constrained environments.

\section{Discussions}
\label{sec::discussions}

While many discussion points from The 1st BARN Challenge~\cite{xiao2022autonomous} are still valid this year, we discuss new findings and lessons from The 2nd BARN Challenge and point out promising future research directions to push the boundaries of efficient mobile robot navigation in highly constrained spaces. 

\subsection{More participants from industry (hybrid academia-industry teams)}
One interesting change in the 2nd year of The BARN Challenge is the participation from robotics industry. Compared to last year's competition, in which all participants were from universities worldwide, five out of the ten teams who participated in The 2nd BARN Challenge were from industry or hybrid academia-industry teams (KU Leuven with Flanders Make, INVENTEC, University of Almeria with Teladoc Health Inc., Staxel, and Lumnonicity from JIO AICOE). Notably all three winning teams have industry ties. 

The significant increase in industry participation indicates that the problem The BARN Challenge aims to solve is of significant interest to real-world robotics manufacturers, providers, and users. As discussed in last year's report~\cite{xiao2022autonomous}, even very experienced robotics researchers in academia may have the impression that such an ostensibly simple problem has already been solved. However, the fact that robotics industry is still working on such a problem and using The BARN Challenge as a testbed for their methods suggests that they still do not have a satisfactory solution for this problem in the real world. Therefore, the organizers suggest academic researchers to consider real-world problems the robotics industry faces in order to realize technology transfer from academia to industry. The 1st and 3rd place winners, KUL+FM and University of Almeria (with Teladoc) are very good examples of applying state-of-the-art academic research to real-world robotics industry and identifying real-world problems from industry to solve together with academia.

\subsection{Simulation performance approaching the current upper bound}
Another observation from the simulation qualifier is that the simulation performance has approached the current metric's upper bound, i.e., $0.25$. The upper and lower bound is determined by the two hyper-parameters $4$ and $8$ in the evaluation metric to  assure navigating extremely quickly or slowly in easy or difficult environments respectively won’t disproportionally scale the score. Approaching $0.25$ means most of the navigation trials in the BARN environments can be successfully finished within four times of the optimal traversal time, even in the most difficult ones with very dense obstacles. Therefore, the organizers will change these hyper-parameters in next year's challenge to increase the upper and lower bound to encourage the teams to score higher by achieving less traversal time than four times of the optimal time, without colliding with any obstacle.

\subsection{The first team that finished all physical courses}
KUL+FM is the first team in The BARN Challenge history that was able to successfully navigate the Jackal through all three physical obstacle courses in all nine final counted trials. However, they did fail three out of the total 15 timed trials in obstacle courses 1 and 3 as shown in Tab.~\ref{tab::physical_results}. While the failure trial in obstacle course 1 was likely due to the need of initial parameter fine-tuning to fit the physical environments, the two failure trials in the last obstacle course were both due to the desire to push on faster navigation speed. Considering that the physical obstacle courses constructed this year were qualitatively more difficult than last year's, all three winning teams' stable performance in the first two obstacle courses and KUL+FM's all nine successful trials in all three courses suggest that the physical performance of the navigation stacks have been improved compared to last year.

\subsection{Final ranking still decided by success rate, not speed}
Despite the improved performance and more confident deployment experience (e.g., less struggling to navigate to the goal and to avoid obstacles, but more focus on fine-tuning for robustness and speed) observed by the organizers, completely collision-free navigation is still out of reach for most teams regardless of speed. Therefore the final ranking in the physical finals is still decided by success rate, not navigation speed. Although the 2nd place winner, INVENTEC, had a chance to win the competition before they started the 3rd obstacle course by pushing on reducing traversal time, they eventually failed all five attempts in the last course. So the community is still waiting for the first BARN Challenge, in which the final ranking is determined by traversal speed in highly constrained obstacle environments, after the success rate has been guaranteed to be $100\%$. 

\subsection{Less obvious sim-to-real gap}
Compared to last year when the winning approach in simulation suffered from significant collisions in the real world~\cite{xiao2022autonomous}, the rankings in Tab.~\ref{tab::sim_results} and~\ref{tab::physical_results} suggest an decreasing sim-to-real gap between the simulation qualifier and physical finals this year. Based on the approaches taken by the teams, it is possible that such a smaller sim-to-real gap is caused by the decreased usage on learning-based navigation methods~\cite{xiao2022motion}. Despite the popularity of using machine learning to address visual inputs~\cite{kahn2021badgr, voilaharesh, sikand2022visual}, off-road conditions~\cite{pan2020imitation, kahn2021badgr, xiao2021learning, karnan2022vi, atreya2022high, datar2023toward, datar2023learning}, social contexts~\cite{mirsky2021prevention, mavrogiannis2021core, scand, xiao2022learning, nguyen2023toward}, and multi-robot navigation~\cite{long2018towards, chen2017decentralized}, only one team, INVENTEC (except Temple TRAIL who did not compete in the physical finals), used a learning-based method to navigate in highly constrained obstacle environments in The 2nd BARN Challenge. Their approach is mostly based on the Learning from Hallucination~\cite{xiao2021toward, xiao2021agile, wang2021agile, park2023learning} paradigm, especially the latest LfLH~\cite{wang2021agile} approach (one of the baselines provided by the competition organizers). Furthermore, INVENTEC designed sophisticated recovery behaviors to address real-world scenarios where the learning approach did not work well (see detailed discussions below), which also helped to reduce the sim-to-real gap.  

\subsection{Importance of good recovery behaviors to deploy end-to-end learning-based systems in the real world}
During The 1st BARN Challenge last year, the end-to-end learning approach trained by Deep Reinforcement Learning by Temple TRAIL~\cite{xie2023drl} experienced significant sim-to-real gap because the training was conducted in simulation on a different, smaller robot platform (Turtlebot2) and there was no recovery behavior. This year, Temple TRAIL was not able to compete in-person and the system deployed by the organizers on behalf of the team did not perform well without fine-tuning. As the other team that adopted machine learning for their navigation system,  INVENTEC used the LfLH~\cite{wang2021agile} policy trained using simulated data, but also devised a set of recovery behaviors to address real-world scenarios where the learning approach did not work well. By developing good recovery behaviors to complement an end-to-end learned motion policy, INVENTEC's approach outperformed the original LfLH approach, which is assisted only by a set of very simple recovery behaviors, and achieved very good sim-to-real transfer. This observation suggests the potential of end-to-end learning approaches for navigation when being augmented by a sophisticated mechanism to complement learning during out-of-distribution real-world scenarios~\cite{liu2021lifelong, xu2022learning}.

\subsection{Tuning still necessary for all classical systems facing different obstacle environments}
Similar to last year's observation, the original intention of ``out-of-the-box'' deployment of the navigation systems submitted to the simulation qualifier directly in the physical finals was still impossible for all the teams: all teams that used classical systems had to extensively fine-tune their navigation stack, while INVENTEC, the only team that used a learning-based approach, fine-tuned their recovery behaviors, the classical part of their system. Although slight fine-tuning to adapt a system from simulation to the real world is reasonable, all teams needed to fine-tune their navigation systems before competing in all three obstacle courses. It is unclear whether a single navigation system configuration that works for all obstacle courses exists or not. The organizers suggest the community to keep such an intention in mind when developing their navigation systems, because the option of fine-tuning for every deployment scenario is not possible, especially when the goal is to deploy autonomous mobile robots at scale in the wild. One single navigation system configuration or parameterization to address all possible scenarios is certainly ideal, but another promising approach is autonomous parameter tuning to adapt to different obstacle configurations around the robot~\cite{xiao2020appld,  wang2021appli, wang2021apple, xu2021applr, ma2021navtuner, xiao2022appl}. 

\subsection{Interest from outside the robotics community}
Surprisingly, The 2nd BARN Challenge has even raised interest from outside the robotics community. For example, Harald Carlens, a Machine Learning researcher who runs \url{https://mlcontests.com}, which lists machine learning competitions, has reported The 1st BARN Challenge after ICRA 2022 in ``The State of Competitive Machine Learning''~\cite{mlcontests}. Harald Carlens was able to attend and observe The 2nd BARN Challenge in-person in London and had long discussions with the organizers regarding The BARN Challenge and other Machine Learning competitions. Another Artificial Intelligence student researcher from RWTH Aachen University, Malte Schwerin, reached out after The 2nd BARN Challenge and requested raw results of the simulation qualifier to further study ranking methods for algorithm competitions. A professor in robotics and psychology from Graz University of Technology in Austria, Dr. Gerald Steinbauer-Wagner, reached out during the physical finals and was interested in assessing robotic capabilities using methods from psychology, i.e., test theory, which had been applied for centuries to asses the latent capabilities of people, e.g., intelligent and PISA test. This well-grounded theory allows to create objective, reliable, and valid tests for skills, even for nondeterministic domains. Dr. Steinbauer-Wagner's research group is currently working on analyzing the BARN simulation runs via Item Response Theory~\cite{cantrell1997item} and One-Parameter Rasch Model~\cite{hambleton1985principles}. Such a trend indicates that benchmarking robotic navigation capabilities is not only of interest for the sake of improving autonomous navigation for roboticists alone, but also has broader impacts on the study of machine learning, artificial intelligence, test theory, psychology, and potentially other disciplines. 

\section{CONCLUSIONS}
\label{sec::conclusion}
The results of The 2nd BARN Challenge at ICRA 2023 suggest that the mobile robot community has been making steady progress on autonomous metric ground navigation, an ostensibly simple but largely unsolved problem. The compeition has doubled its size, attracted many participants from robotics industry, approached the performance upper bound in the simulation qualifier, had the first winning team that finished all physical trials, experienced a smaller sim-to-real gap between the simulation qualifier and physical finals, and attracted interest from outside the robotics community. However, work remains to be done in order to guarantee collision avoidance and to reduce the need of system tuning to efficiently navigate in different obstacle environments. 

\bibliographystyle{IEEEtran}
\bibliography{IEEEabrv,references}
\end{document}